\pdfoutput=1

\documentclass[11pt]{article}

\usepackage[final]{acl}
\usepackage{multirow}
\usepackage{times}
\usepackage{latexsym}
\usepackage{algorithm,algorithmic}

\usepackage[mathscr]{euscript}
\usepackage{times}
\usepackage{latexsym}
\usepackage{booktabs}
\usepackage{graphicx}
\usepackage{enumitem}
\usepackage{amsmath}
\usepackage{amsfonts}
\usepackage{listings}
\usepackage{fvextra}
\usepackage{fancyvrb} 
\usepackage{framed}   
\usepackage{mdframed} 
\usepackage{fvextra}
\usepackage{listings}
\usepackage{fancyvrb}
\usepackage{framed}  
\usepackage[T1]{fontenc}

\usepackage[utf8]{inputenc}

\usepackage{microtype}

\usepackage{inconsolata}

\usepackage{graphicx}

\definecolor{citecolor}{HTML}{2980b9}
\definecolor{linkcolor}{HTML}{c0392b}
\definecolor{mycolor_blue}{HTML}{E7EFFA}
\definecolor{mycolor_green}{HTML}{E6F8E0}

\title{Cognitive-Level Adaptive Generation via Capability-Aware Retrieval and Style Adaptation}

\author{Qingsong Wang$^{1}$,
Tao Wu$^{1}$,
Wang Lin$^{1}$,
Yueying Feng$^{1}$\\ 
\textbf{Gongsheng Yuan}$^{1}$, 
\textbf{Chang Yao}$^{1}$ ,
\textbf{Jingyuan Chen}$^{1}$\thanks{~Corresponding author.} \\
$^{1}$ Zhejiang University\\
\texttt{\small wqsong@zju.edu.cn}\\ \texttt{\small jingyuanchen@zju.edu.cn}
}

\begin{document}
\maketitle

\begin{abstract}

Large Language Models (LLMs) have demonstrated strong performance in open-ended generation tasks. However, they often struggle to adapt content to users with differing cognitive capacities, leading to a phenomenon we term cognitive misalignment. This issue arises in two forms: knowledge-level misalignment, where content is too complex or too simplistic relative to user understanding, and presentation style misalignment, where the structure or tone hinders effective comprehension. To address these challenges, we propose the Cognitive-Level Alignment Framework (CLAF), a general-purpose generation framework that aligns both knowledge complexity and presentation style with user cognition. CLAF integrates a capability-aware retrieval module based on a hierarchical knowledge graph and a style optimization module guided by Bloom’s taxonomy and preference learning. Additionally, a knowledge-controllable generation component ensures consistency and relevance throughout the output. To support training and evaluation, we construct Scale, a cognitively annotated dataset containing responses at multiple comprehension levels per query. Empirical results show that CLAF enhances the adaptability and informativeness of LLM outputs across a range of user profiles, offering a robust solution to cognitive-level alignment in real-world applications. Code and dataset are available at:  \href{https://anonymous.4open.science/r/lg-90A8}{https://anonymous.4open.science/r/lg-90A8}

\end{abstract}

\begin{figure}[t]
  \includegraphics[width=1\linewidth]{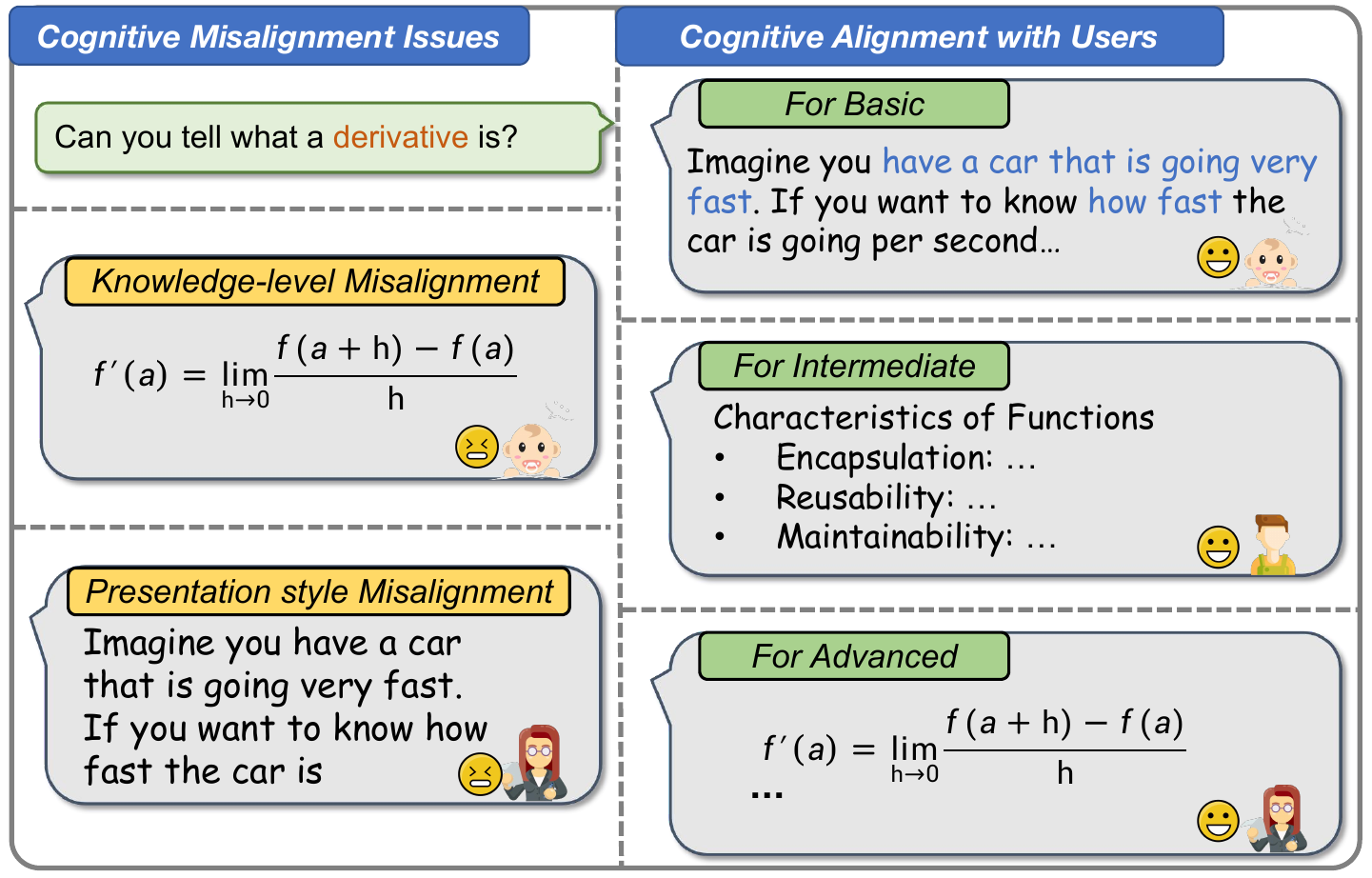}
  \vspace{-0.5cm}
  \caption {\textbf{Comparison of Cognitive Misalignment and Alignment in LLMs.} The left side illustrates cognitive misalignment, making the content difficult to understand or boring. In contrast, the right side demonstrates correct cognitive alignment, where everyone receives a suitable response.}
  \vspace{-0.5cm}
    \label{intro}
\end{figure}

\section{Introduction}

Large Language Models (LLMs) have shown remarkable capabilities in the field of education~\cite{lin2025non,wang2025omni,huang2025autogeo}. Building upon these strengths, LLMs also enable personalized education by adapting their responses to individual learners’ needs and communication styles across diverse educational and professional scenarios
~\cite{han2025contrastive,wu2025embracing,kwon-etal-2024-biped,feng2024fedpam}. 
Central to the effectiveness of these models is their capacity to adapt responses to users' varying levels of cognitive ability---ensuring that generated content is not only accurate but also aligned with the users' capacity for comprehension~\cite{poole2024llm,wu2025personalized}. However, current LLMs frequently fail to achieve this alignment, resulting in a phenomenon we term \textbf{cognitive misalignment}, which impairs the instructional efficacy of model outputs~\cite{liu2024progressively, rooein2023know}.

Cognitive misalignment manifests in two primary forms, as illustrated in Figure~\ref{intro}. The first is \textit{knowledge-level misalignment}~\cite{he2024psychometric}, where the complexity of the content exceeds or underestimates users' cognitive capacity. For instance, a novice user may receive explanations embedded with technical jargon, while an expert user may be presented with overly simplistic content, leading to disengagement or frustration. The second is \textit{presentation style misalignment}~\cite{sonkar2024pedagogical}, which arises when the communicative approach fails to align with users' instructional needs. Similar to how educators tailor pedagogical strategies to users, LLMs should ideally adapt their rhetorical style, explanatory granularity, and instructional scaffolding accordingly.
However, many models default to rigid stylistic patterns, resulting in suboptimal educational interaction for users across varied cognitive levels.

While recent efforts in personalized text generation have made notable progress~\cite{liu2024llms+,singh2024personal}, they generally fall short of addressing these two dimensions of cognitive misalignment. Most approaches focus on user interests or interaction history, with limited consideration of users' cognitive ability. As a result, such approaches often capture \textbf{what} users are curious about, but not \textbf{how} that information should be \textbf{structured for effective comprehension}. Similarly, presentation style adaptation efforts typically depend on extensive personalization data, which is unavailable in many real-world contexts where users are represented by coarse-grained profiles (\textit{e.g.}, ``middle school student'' or ``domain expert'')~\cite{liu2024inter}. These abstractions provide insufficient granularity for pedagogically appropriate adaptation, resulting in uniform outputs that inadequately serve diverse cognitive needs.

To address these limitations, we propose the \textbf{C}ognitive-\textbf{L}evel \textbf{A}lignment \textbf{F}ramework (CLAF), a novel architecture designed to jointly align both content complexity and instructional style with users' cognitive level. 
Grounded in principles from educational psychology, particularly Vygotsky’s Zone of Proximal Development (ZPD)~\cite{nogueira2001vygotsky} and Bloom’s Taxonomy of Educational Objectives~\cite{huitt2011bloom}, CLAF integrates cognitive theory with LLM capabilities to systematically address both dimensions of cognitive misalignment.
Figure 4 illustrates the overall architecture of CLAF. 

To mitigate knowledge-level misalignment, CLAF employs a \textbf{capability-aware retrieval} module inspired by ZPD. This module constructs a hierarchical knowledge graph organized by cognitive complexity, allowing for the retrieval of content that is optimally challenging yet comprehensible for the user's developmental stage. 
To address presentation style misalignment, we introduce an \textbf{adaptive language style optimization} module informed by Bloom's taxonomy and reinforced via human preference optimization. This module adjusts the explanatory tone, rhetorical structure, and pedagogical strategies based on the user's cognitive capabilities (\textit{e.g.}, remembering, understanding, and applying), thereby supporting the the generation of content that is both personalized and instructionally coherent. Moreover, to ensure alignment between retrieved instructional content and generated text, CLAF incorporates a \textbf{knowledge controllable generation} mechanism that constrains latent representations during decoding, thereby preserving coherence and pedagogical relevance.

To support empirical validation, we construct a novel dataset, \textbf{Scale}, comprising responses at three distinct cognitive levels per question---designed to reflect Bloom's taxonomy and instructional design principles. This dataset functions as both a training signal and an evaluation benchmark for cognitive-level adaptive generation.

\begin{figure}[t]
  \includegraphics[width=1\linewidth]{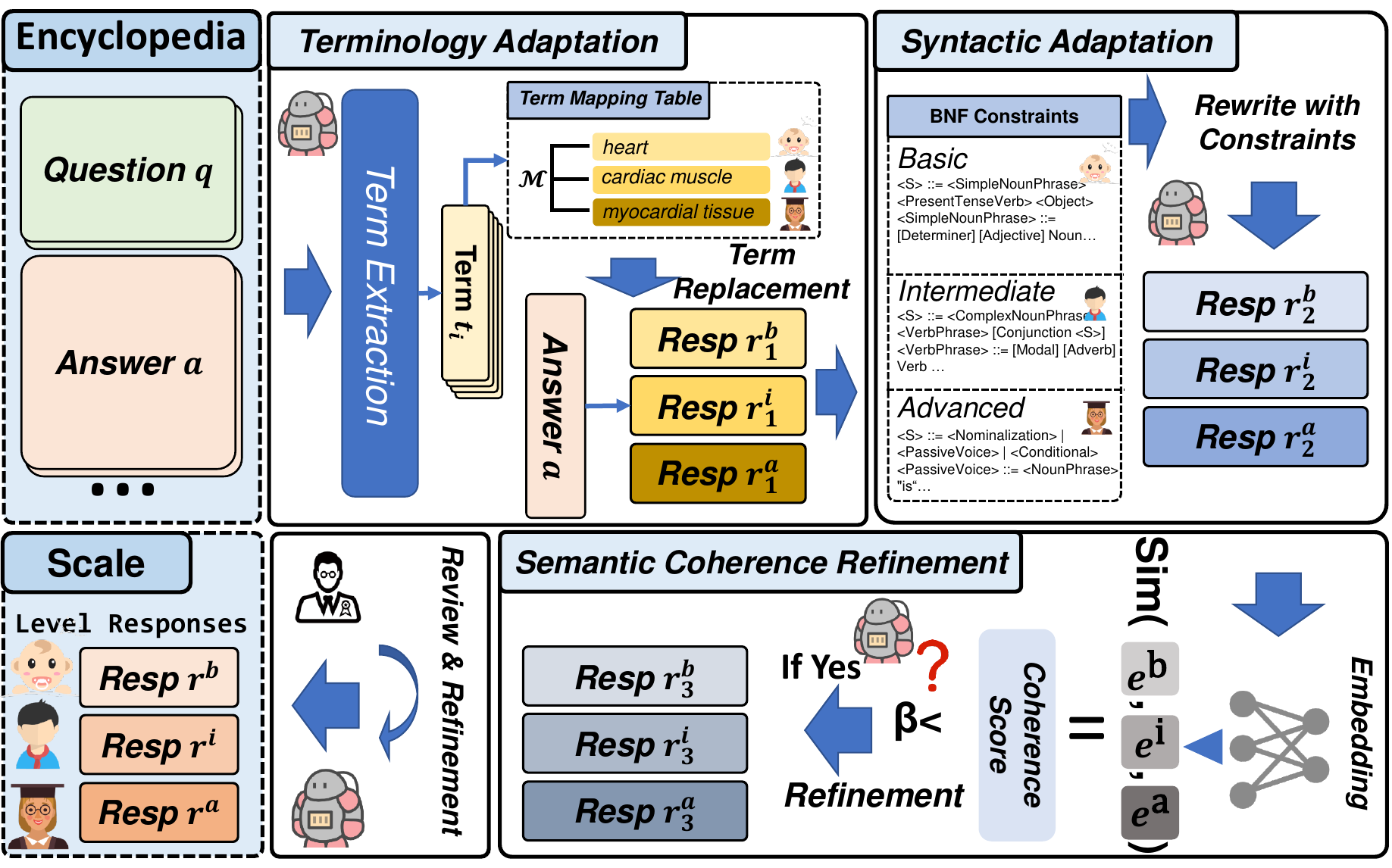} 
\vspace{-0.6cm}
  \caption {\textbf{Construction Pipeline of Scale.} Scale is built from encyclopedia question-answer pairs, forming a dataset where each question is associated with three responses, each customized for a different knowledge level.}
    \label{dataset}
\vspace{-0.4cm}
\end{figure}

Our contributions are summarized as follows:
\begin{itemize}[itemsep=0pt,topsep=0pt,parsep=0pt,leftmargin=*]
\item We identify and formally define \textbf{cognitive misalignment} as a core limitation in current LLM-based systems, framed through cognitive development theory.
\item We propose \textbf{CLAF}, a cognitively grounded generation framework that jointly optimizes content retrieval, linguistic adaptation, and instructional fidelity.
\item We construct \textbf{Scale}, a cognitively annotated dataset that enables systematic training and evaluation of cognitive-level alignment in LLMs.
\end{itemize}

\section{Related Work}
\subsection{Personalized Large Language Model}

The capabilities of LLMs can be leveraged for personalized teaching. \citep{park2024empowering,neshaei2024towards,tang2024morpheus} utilize students' historical conversations and personal information to model students. \citep{hu2024foke} integrates knowledge graphs and prompt engineering into LLMs. \citep{deng2023towards,chen2023gptutor,li2024learning} select the next learning goal for students based on analysis by LLMs. All these works use students' historical information to model them, aiming to achieve personalized education.

While personalized education excels at modeling individual students, it struggles with group. They continuously analyzes individual learning trajectories, such as knowledge retention patterns and thinking path deviations. In contrast, modeling group cognitive level focuses on group cross-sectional data, categorizing students into homogeneous groups based on predefined proficiency indicators like learning stages. It then designs standardized teaching programs by identifying common features at each cognitive level.

\subsection{Controlled text generation}

Significant advancements have been made in controllable text generation methods now. DisCup~\cite{zhang2022discup} enhances control by introducing attribute discriminators during training and optimizing control cues through anti-likelihood training. RMT~\cite{zhang2023controllable} adopts residual learning and cross-attention mechanisms to achieve text generation control and seamlessly integrates with existing LLMs to enable continuous control. REI~\cite{zheng2023toward} uses instructions inspired by regular expressions to control text generation through linguistic constraints. In addition to the above methods that require training, ICV~\cite{liu2023context} learns control-related vectors through contextual example text, effectively enhancing controllable text generation (CTG). MacLaSa~\cite{ding2023maclasa} uses variational autoencoders (VAE) to map text to a compact latent space and applies ordinary differential equation (ODE) sampling methods to control multiple attributes.

\begin{figure*}[t]
  \centering
  \includegraphics[width=1\linewidth]{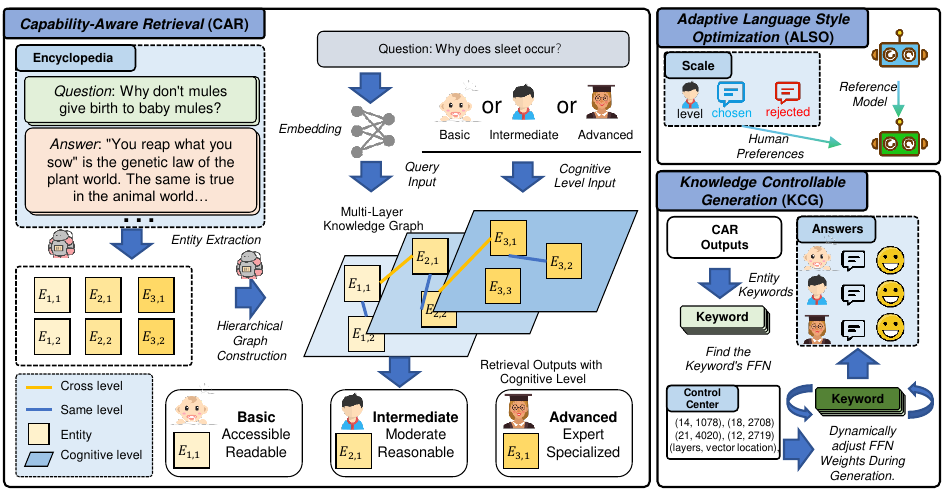} 
    \vspace{-1cm}
  \caption {\textbf{Overview of the CLAF Framework.} The framework consists of three modules: Capability-Aware Retrieval, Adaptive Language Style Optimization, and Knowledge Controllable Generation.}

    \vspace{-0.3cm}
    \label{framework_fig}
\end{figure*}

\section{Dataset Curation}
Effective research on cognitively aligned text generation needs datasets with two key features: (1) clearly defined cognitive levels, and (2) responses that share the same meaning but vary in depth of explanation and language complexity. Existing datasets on personalization~\cite{liu2024personality,zheng2020pre,shen2024pmg} often focus on user traits or history, but they lack structured cognitive levels and controlled variation in how the answers are written.
To address this, we introduce the \textbf{Scale}, designed to support controlled generation across different cognitive levels. Each item in Scale includes a question and three aligned answers: one each at the \textit{basic}, \textit{intermediate}, and \textit{advanced} levels. These levels are based on cognitive development theory (see Appendix~\ref{appendix:cognitive_detial}), and differ in how ideas are explained while keeping the core meaning intact. To check the quality of the answers, we use a multi-step human evaluation. Further information about human evaluation can be found in Appendix~\ref{appendix:data_detail}.
\subsection{Metadata Collection and Domain Scope}
\label{metadata_collection}
We build the core QA pairs using reliable and informative encyclopedic sources, covering topics like science, nature, biology, and cosmology. These areas are chosen for their broad appeal and their suitability for explaining topics at multiple levels of detail. The questions include types like definitions, explanations, and cause-effect reasoning, offering a range of reasoning modes.
To test generalization, we also create a separate test-only set based on Chinese classical poetry, taken from national college entrance exam materials. These expert-written questions help evaluate performance across both domain and language. This part is only used in testing and not seen during training.
\subsection{Data Construction Pipeline}
As shown in Figure~\ref{dataset}, we build the ``Scale'' in three steps to adapt it to users at different levels:

\textbf{Step 1: Terminology Adaptation.} Since word choice is key to abstraction, we extract important terms from each original answer. Using LLMs, we generate versions of these terms for each level (\textit{basic}, \textit{intermediate}, \textit{advanced}). Experts then review and confirm these mappings, which are used to create different wordings for each answer: $\mathcal{R}_1 = \{{r_1}^b, r_1^i, r_1^a\}$.

\textbf{Step 2: Syntactic Adaptation.} Beyond words, sentence structure also affects comprehension. We define templates based on Backus-Naur Form (BNF)~\cite{mccracken2003backus} and teaching guidelines. Basic answers use short, direct sentences. Intermediate ones use compound clauses and general ideas. Advanced answers include abstract sentence patterns and more layered grammar. We apply these rules using prompts to generate syntactically distinct versions: $\mathcal{R}_2 = \{r_2^b, r_2^i, r_2^a\}$.We also match the responses with typical learning patterns: basic answers use familiar examples and surface facts; intermediate ones add general explanations and causes; advanced ones include inferential thinking and disciplinary concepts.

\textbf{Step 3: Semantic Coherence Verification.} To ensure all answers still mean the same thing, we use vector-based similarity checks. If a rewritten answer is too different in meaning, it is revised. This yields the final triple-set: $\mathcal{R}_3 = \{r_3^b, r_3^i, r_3^a\}$, preserving meaning while varying complexity.

\subsection{Dataset Summary}
The final Scale contains 593 question-and-answer entries, each with three levels of response. In addition to enabling controlled text generation, Scale supports consistent evaluation of language models’ ability to change abstraction and tone while keeping meaning stable. Our modular pipeline allows future extensions to new topics and levels. The extra test set in classical Chinese literature provides a challenge for models in cross-lingual and content-rich understanding tasks.

\begin{table*}[th!]
\begin{center}
  \scalebox{0.8}{
\begin{tabular}{lcccccccccc}

\toprule
\multirow{2}{*}{Model} & \multicolumn{2}{c}{Flesch Kin} & \multicolumn{2}{c}{Gunning Fog} & \multicolumn{2}{c}{SMOG} & \multicolumn{4}{c}{Match Level} \\ 
\cmidrule(lr){2-3}  \cmidrule(lr){4-5} \cmidrule(lr){6-7} \cmidrule(lr){8-11}
                       & Bas.$\downarrow$  & Adv. $\uparrow$   & Bas.$\downarrow$ & Adv. $\uparrow$   & Bas.$\downarrow$  & Adv.$\uparrow$   & Bas.  $\uparrow$   & Int. $\uparrow$   & Adv. $\uparrow$   & Avg. $\uparrow$ \\ 
\midrule
\multicolumn{11}{c}{\textit{Closed-source LLMs}}                                                                                                                     \\
    GPT-4o              & 6.97           & 14.19           & 8.31           & 16.13           & 8.38            & 15.94          & 79.85            & 85.30           & 84.74           & 83.30 \\
    GPT-4o-FS           & 6.55           & 14.97           & 7.89           & 17.10           & 7.85            & 16.19          & 89.72            & 85.69           & 90.52           & 88.65 \\
    Gemini-1.5          & 6.77           & 13.10           & 8.07           & 14.19           & 8.54            & 14.48          & 83.57            & 74.50           & 82.73           & 80.27 \\
    Gemini-1.5-FS       & 6.17           & 13.21           & 7.49           & 14.10           & 8.08            & 14.47          & 80.07            & 84.36           & 82.70           & 82.38 \\
    Claude-3.5          & 7.36           & 15.66           & 8.54           & 16.53           & 9.21            & 16.43          & 73.75            & 85.48           & 73.86           & 77.70 \\
    Claude-3.5-FS       & 7.15           & 15.88           & 8.39           & 16.85           & 8.69            & 16.60          & 78.33            & 84.90           & 77.15           & 80.13 \\
    Qwen-Plus           & 6.60           & 13.79           & 7.87           & 14.56           & 8.06            & 15.14          & 67.91            & 89.07           & 70.40           & 75.79 \\
    Qwen-Plus-FS        & 6.47           & 13.80           & 7.91           & 14.98           & 8.34            & 15.21          & 77.17            & 87.85           & 79.45           & 81.49 \\
\midrule           
\multicolumn{11}{c}{\textit{Qwen-2.5-3B-Instruct}}  \\           
Few-Shot                & 7.17           & 12.99           & 8.44           & {14.45}         & 8.13            & 14.24            65.79           & 87.54           & 68.47           & 73.93 \\
SFT                     & 6.91           & {13.29}         & 8.26           & 14.69           & 8.32            & 14.38           & 78.33           & 82.80           & 79.37           & 80.17 \\
CLAF(ours)                    & {6.69}         & 12.80           & {8.10}         & 14.06           & 8.17            & {14.43}         & 76.43           & 85.94         & 81.15           & 81.17 \\
\midrule  
\multicolumn{11}{c}{\textit{Qwen-2.5-7B-Instruct}}   \\  
Few-Shot                & 6.80           & 13.01           & 8.08           & 13.85           & 8.74            & 14.71           & 76.01           & 86.93           & 75.07           & 79.34 \\
SFT                     & 6.37           & 13.64           & 7.72           & 14.58           & 8.06            & \textbf{15.23}  & 79.00           & 81.15           & 77.55           & 79.23 \\
CLAF(ours)                    & \textbf{5.81}  & 13.47           & \textbf{7.16}  & 14.50           & \textbf{8.02}   & 15.04           & 78.01           & {87.63}  & 81.63           & 82.42 \\

\midrule  
\multicolumn{11}{c}{\textit{Llama-3.1-8B-Instruct}}    \\  
  
Few-Shot                & 7.17           & 13.32           & 8.51           & 13.70           & 9.09            & 14.84           & 26.35           & \textbf{92.80}           & 28.90            & 49.35 \\
SFT                     & 6.60           & 13.92           & 8.30           & 16.21           & 8.37            & 12.90           & 85.53           & 78.15           & 78.78            & 80.82 \\
CLAF(ours)                    & 6.25           & \textbf{13.78}  & 8.19           & \textbf{16.38}  & 8.14            & 14.22           & \textbf{90.75}  & {86.30}         & \textbf{90.87}   & \textbf{89.31} \\
\bottomrule
\end{tabular}
}

  \caption{\label{table:main}\textbf{Experimental Comparison of CLAF Against Other Baseline Models.} The results validate the effectiveness of the proposed framework.}
\end{center}
\end{table*}

\vspace{-0.1cm}

\section{Methodology}

This paper presents an innovative framework by aligning generated content with distinct cognitive levels. The framework adapts the scope of knowledge, language styles, and teaching strategies in response to the user's cognitive boundaries. The proposed Cognitive Level Alignment Framework (\textbf{CLAF}) consists of three components: 1) Capability-Aware Retrieval, which delivers relevant knowledge tailored to various cognitive levels by retrieving content situated within the user’s proximal development zone, as inspired by ZPD; 2) Adaptive Language Style Optimization (ALSO), which allows the model to employ language styles appropriate for different users by adapting tone and pedagogical strategy based on Bloom’s taxonomy; and 3) KCG, which dynamically adjusts the scope of knowledge and ensures the output remains faithful to the retrieved content. The overview of the framework is illustrated in Figure~\ref{framework_fig}.

\subsection{Capability-Aware Retrieval}

The first step toward cognitive alignment is ensuring the knowledge matches the user’s cognitive level. Inspired by~\cite{jin2025recognize,feng20243,wang2023weakly} ,CAR achieves it by building a hierarchical knowledge graph derived from educational materials, where each node represents an atomic concept labeled with a cognitive tier \( l \in \{0,1,2\} \), corresponding to basic, intermediate, and advanced levels, roughly aligned with Bloom’s taxonomy (\textit{e.g.}, Remember/Understand, Apply/Analyze, Evaluate/Create). Relations among nodes encode prerequisite chains, logical dependencies, and topic proximity.

This structure enables CLAF to perform Bloom-informed, ZPD-aware retrieval. For a user at level $c$, CAR traverses the graph to extract a subgraph $K^{(k)}_c$, constrained to nodes with $l \leq c$, and a depth $d$ that increases with $c$ . As such, beginners are exposed to foundational content, while advanced users receive broader and deeper knowledge. The retrieved concepts serve as inputs for downstream modules. Full retrieval procedures are detailed in Algorithm~\ref{algo:hie} and Appendix~\ref{algo:hie2}.

\subsection{Adaptive Language Style Optimization}
To further refine the alignment of content with the user's cognitive level, we introduce the ALSO module. This module leverages Direct Preference Optimization (DPO)~\cite{rafailov2023direct} to tailor the language style and complexity according to the user's stage. By using Scale, ALSO adapts the style of large language models (LLMs) to match cognitive requirements, dynamically adjusting aspects such as term difficulty, sentence structure, and pedagogical approach.

Unlike static prompt-based strategies, our approach continuously adapts to the user's needs. The DPO framework fine-tunes the model by maximizing the expected reward of the output style while minimizing its divergence from a reference model. The optimization is expressed as:
\begin{align}
\max_{\pi_\theta}\; & \mathbb{E}_{x \sim \mathcal{D},\, y \sim \pi_\theta(y | x)} \left[ r_\phi(x, y) \right] \nonumber \\
& - \beta D_{\text{KL}}\left(\pi_\theta(y | x) \,\|\, \pi_{\text{ref}}(y | x)\right)
\end{align}
where \(\pi_\theta(y | x)\) represents the model's output distribution, \(r_\phi(x, y)\) is the reward function, \(\beta\) controls the trade-off between reward and divergence, and \(D_{\text{KL}}\) is the Kullback-Leibler divergence between the model and the reference model.

\noindent\textbf{Cognitive-level Adaptation} The module tailors responses to user' capabilities through three-tier adaptation inspired by Bloom’s taxonomy of cognitive objectives: For \textit{basic-level} users, it simplifies concepts using fundamental terminology, analogies, and clear explanations aimed at fostering lower-order cognitive processes such as remembering and understanding. The output distribution $\pi_\theta(y | x)$ is optimized via reward modeling to prioritize accessibility. \textit{Intermediate-level} users receive balanced explanations that integrate foundational knowledge with logical reasoning and contextual examples, supporting mid-level cognitive goals such as applying and analyzing. \textit{Advanced-level} users obtain domain-specific terminology and deductive reasoning aligned with expert-level cognition, aligning with higher-order objectives such as evaluating and creating. The preference mechanism follows:
\begin{align}
\label{eq:dpo}
P(y_w \succ y_l | x) = \sigma(r(x, y_w) - r(x, y_l)),
\end{align}
where $P(y_w \succ y_l | x)$ denotes the probability of output $y_w$ being preferred over $y_l$, $\sigma$ is the logistic function, and $r(\cdot)$ represents the reward model that evaluates cognitive alignment.

\subsection{Knowledge Controllable Generation}
To enhance the consistency between LLMs output and CAR's retrieval content, we incorporate a Knowledge Controllable Generation(KCG) module that enables precise control over the output. This approach, based on prior work \cite{feng-etal-2024-freectrl,hu2021knowledgeable,feng2024unveiling}, allows for adaptive management of output's content by adjusting the weights of vectors in the Feedforward Network (FFN) layers. The module constructs a control center for each token in the model’s vocabulary, influencing the generation of domain-specific content by modifying the FFN vector weights.

The generation process consists of four stages: initialization, monitoring, adaptation, and filtering. In the {initialization stage}, relevant keywords from the CAR module are collected, and their corresponding FFN vectors are identified. The {monitoring stage} evaluates the relevance of each token generated, dynamically adjusting weights to optimize domain-specific alignment. During the {adaptation stage}, the weights of the control centers are modified to guide the model towards generating content that aligns with the desired knowledge scope. The weight adjustment is given by:
\begin{align}
\omega_{a_i}^{t+1} = \lambda \cdot \sigma\left(-(\mu_\omega - \hat{\mu}_{a_i}^t) \cdot l_t\right),
\end{align}
where \(\sigma\) is the sigmoid function, \(\mu_\omega\) is a predefined threshold, and \(\hat{\mu}_{a_i}^t\) represents the cumulative alignment. This dynamic weight adjustment prevents over-specialization by resetting weights when alignment exceeds thresholds. Finally, in the {filtering stage}, thresholds are applied to ensure the quality and relevance of the generated content.

The KCG enables precise control over the generation process, enhancing the relevance of output content to knowledge retrieved by the CAR.

\vspace{-0.1cm}
\section{Experiments}

\subsection{Experimental Setups}

\noindent\textbf{Baselines.} We compare CLAF with the open-source models LLaMA 3.1-8B-Instruct~\cite{touvron2023llama} and Qwen-2.5-7B-Instruct~\cite{qwen},Qwen-2.5-3B-Instruct, as well as the closed-source models ChatGPT-4o~\cite{achiam2023gpt}, Gemini 1.5~\cite{team2023gemini}, Qwen-Plus~\cite{qwen}, and Claude 3.5~\cite{anthropic2024claude}.

\noindent\textbf{Metrics.} We assess text readability and complexity using Flesch-Kincaid Grade Level(FK)~\cite{solnyshkina2017evaluating}, Gunning Fog Index~\cite{gunning1969fog}, and SMOG Index~\cite{mc1969smog}. Cognitive hierarchical alignment is evaluated using GPT-o1. See prompts in Appendix \ref{appendix:metrics}.

\begin{figure}[t]
  \includegraphics[width=1\linewidth]{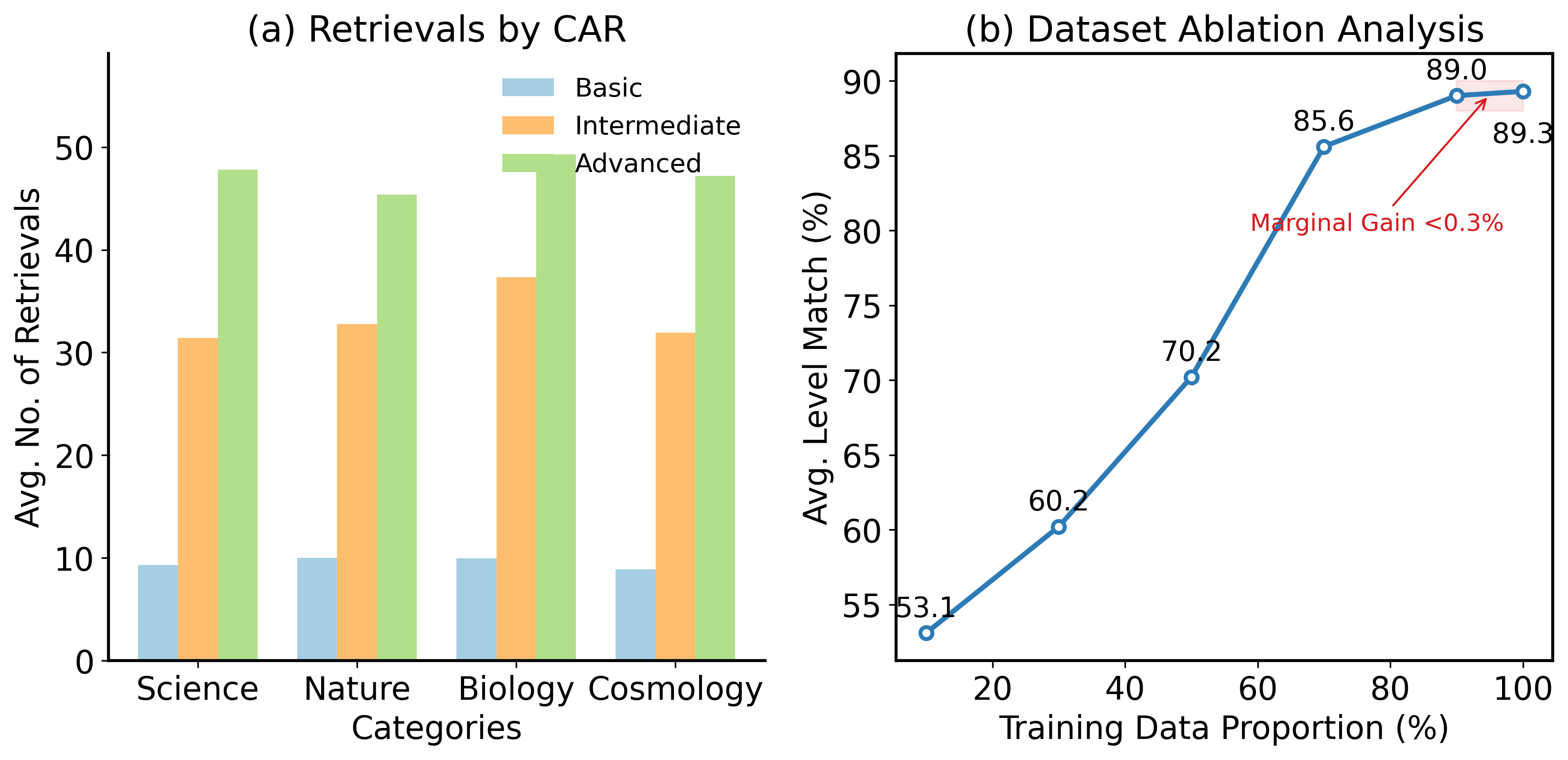} 

  \caption {(a) Number of retrievals in the CAR across different knowledge levels and question types. (b) Results of the Scale-size experiment, indicating that the current dataset volume is sufficient.}

    \label{data_ablation}
\end{figure}

\begin{table}[t]
  \centering
  \resizebox{\linewidth}{!}{
  \begin{tabular}{lcccccc}
    \toprule
     \multirow{2}{*}{Model}& \multicolumn{2}{c}{{Flesch Kin}} & \multicolumn{3}{c}{{Level Match}} \\ \cmidrule(lr){2-3}  \cmidrule(lr){4-6}
           & Bas. $\downarrow$ &Adv. $\uparrow$  & Bas. $\uparrow$ &Int. $\uparrow$ &Adv. $\uparrow$\\
\midrule
        CLAF           & \textbf{8.19}   & \textbf{16.38}  & \textbf{90.75}  & \textbf{86.30}  & \textbf{90.87} \\
        - w/o KCG      & 8.24    & 16.36  & 89.99  & 85.75  & 90.22 \\
        - w/o CAR    & 8.51    & 13.70  & 84.78  & 78.35  & 78.83 \\
        - w/o ALSO     & 9.79    & 14.06  & 57.17  & 76.67  & 59.11 \\
    \bottomrule
  \end{tabular}
  }

  \caption{\label{table:ablation}
    \textbf{Ablation Study of the CLAF Framework.} Results demonstrating the effectiveness of each component model within the CLAF framework.
  }
  \vspace{-0.4cm}

\end{table}

\subsection{Results}

\begin{table}[t]
  \centering
  \resizebox{0.9\linewidth}{!}{
  \begin{tabular}{lccc}
    \toprule
     \multirow{2}{*}{Model}&  \multicolumn{3}{c}{{Level Match}} \\ \cmidrule(lr){2-4}  
         & Bas. $\uparrow$ &Int. $\uparrow$ &Adv. $\uparrow$ \\
        \midrule
        
        \multicolumn{4}{c}{\textit{Closed-source LLMs}} \\
        GPT-4o-FS        & 50.01 & 87.81 & 55.96  \\
        Gemini-1.5-FS    & 51.05 & 86.61 & 57.30  \\
        Claude-3.5-FS    & 51.27 & 83.63 & 61.19  \\
        Qwen-Plus-FS     & \textbf{56.78} & \textbf{88.01} & \textbf{61.85}  \\ 
        
        \midrule
        \multicolumn{4}{c}{ \textit{Open-source LLMs}} \\

    QwQ-32B-Preview-FS   & 35.00 & 71.51 & 58.94  \\
        Llama-3.1-8B     & 22.05 & \textbf{84.32} & 38.26  \\
        Llama-3.1-8B-SFT & 55.46 & 79.04 & 57.71  \\
        CLAF(ours)            & \textbf{60.55} & 80.07 & \textbf{64.97}  \\

    \bottomrule
  \end{tabular}}

  \caption{\label{table:prom}
   \textbf{Experimental Results on the Chinese Classical Poetry Appreciation Dataset.} Results demonstrating the strong generalization ability of CLAF across different domains and languages.}
\vspace{-0.4cm}
  
\end{table}

We evaluated various models, including our proposed one, across different cognitive levels, with results in Table \ref{table:main}. The models' outputs were assessed using the Flesch-Kincaid Grade Level, Gunning Fog Index, SMOG, and Level Match to measure precision and appropriateness. "FS" refers to few-shot prompts, and "Supervised Fine-Tuning (SFT)" involves fine-tuning with CLAF.

The Flesch-Kincaid, Gunning Fog, and SMOG indices assess sentence complexity, with higher scores indicating more difficulty. Basic-level users benefit from lower indices for better comprehension, while Advanced-level users benefit from higher indices. Intermediate-level users require a balance between the two for optimal learning.

\begin{figure*}[t]
\centering
  \includegraphics[width=0.85\linewidth]{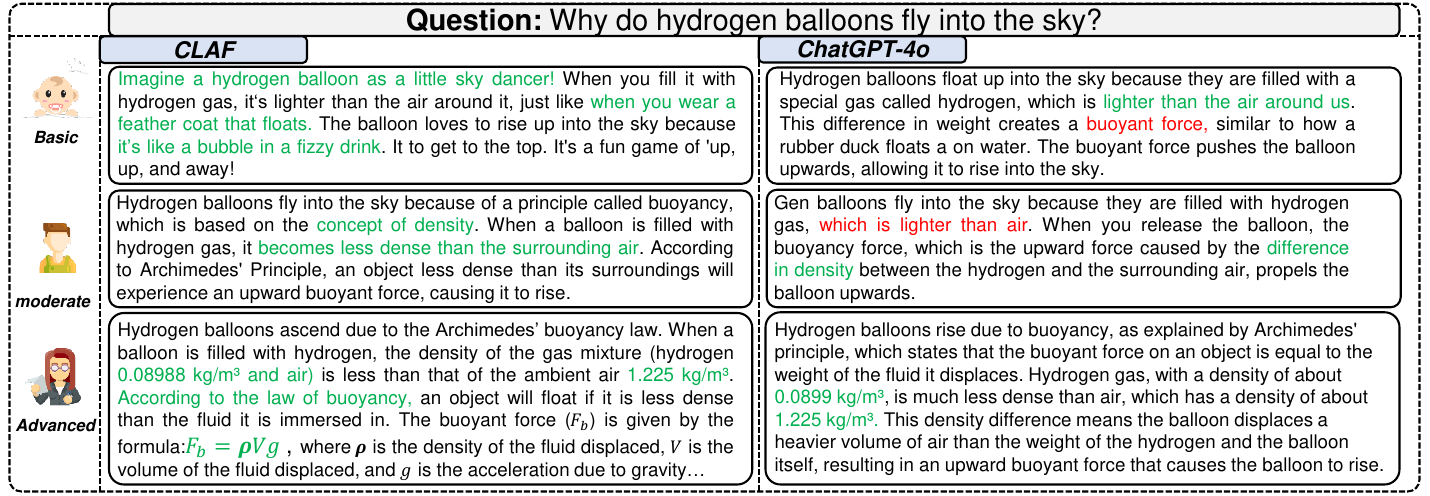} 
\vspace{-0.1cm}

  \caption {\textbf{Case Study.} The results indicate that our CLAF achieves better cognitive level alignment. }
\vspace{-0.2cm}
  
    \label{case}
\end{figure*}

\noindent\textbf{Overall Performance.}The results in Table \ref{table:main} show that CLAF significantly enhances cognitive alignment in text generation. By integrating ALSO with CAR, CLAF improves readability and hierarchical matching rates. It reduces the Flesch-Kincaid score by 5.3\% for basic-level outputs on Llama-3.1-8B and increases the Gunning Fog score by 1.05 points for advanced-level outputs, indicating effective complexity management. The SMOG scores (Bas.=8.02, Int.=12.81, Adv.=15.04) on Qwen-7B surpass Gemini-1.5-FS, validating KCG's role in academic depth modulation. CLAF enables Qwen-7B to achieve an 81.63 advanced matching rate, outperforming Qwen-Plus (70.40). For open-source models, it achieves an 89.31\% average matching rate on Llama-3.1-8B, surpassing Few-Shot and SFT baselines by 40.96 and 8.49 percentage points, respectively. The CAR mechanism boosts advanced generation to a 90.87\% matching rate, outperforming state-of-the-art closed models.

\noindent\textbf{Effectiveness of the Scale.} The Scale demonstrates effectiveness in three ways: (1) In closed-source models, few-shot prompting with Scale improves performance, with GPT-4o-FS achieving a 90.52\% advanced-level match (+5.78\% over zero-shot); (2) For open-source models, full-parameter SFT training with Scale significantly enhances capabilities, with Llama-8B-SFT reaching an 80.82\% average match rate (+63.8\% over the few-shot baseline) and improving basic-level performance from 26.35\% to 85.53\%; (3) Scale identifies catastrophic failure patterns in unadapted models, such as Llama-FewShot's low basic (26.35\%) and advanced (28.90\%) matching rates, and highlights the bias toward intermediate content (92.80\% Int. match), effectively diagnosing LLM bias through contrastive evaluation.

\noindent\textbf{Impact of Model Scaling.} Model scaling experiments show framework adaptability: Qwen-7B improves the average matching rate by 1.25 points (82.42 vs 81.17) over Qwen-3B, with SMOG scores rising from 14.43 to 15.04, indicating larger models better utilize KCG signals. Notably, our method surpasses most closed-source models in advanced matching using only 10-25\% of the parameters of commercial models (Llama-3.1-8B vs Claude-3.5), demonstrating effectiveness under limited computation.
These results validate the tripartite mechanism: 1) ALSO creates granular linguistic representations via DPO; 2) CAR dynamically constrains knowledge boundaries during generation; 3) KCG ensures output content relevance to the question. The framework's multi-objective optimization enables new applications in educational content generation and personalized information delivery with hierarchical adaptation.

\begin{table}[t]
\resizebox{\linewidth}{!}{
\begin{tabular}{lccccc}
\toprule
{CLAF} & FK-Bas.$\downarrow$ & FK-Adv.$\uparrow$ & LM-Bas.$\uparrow$ & LM-Int.$\uparrow$ & LM-Adv.$\uparrow$ \\
\midrule
 0-25\% & \textbf{8.34} & \textbf{14.32} & \textbf{87.69} & \textbf{84.31} & \textbf{86.76} \\
w/o KCG 0-25\% & 8.45 & 13.82 & 84.34 & 84.39 & 80.14 \\

\midrule
 25-50\% & \textbf{8.29} & \textbf{15.97} & \textbf{88.41} & \textbf{85.12} & \textbf{88.09} \\
w/o KCG 25-50\% & 8.33 & 14.88 & 85.43 & 82.46 & 85.35 \\
\bottomrule
\end{tabular}
}
\centering

\caption{\textbf{Ablation Study of the KCG Under Poor Retrieval Conditions}. The results show that KCG brings more significant performance gains when the quality of retrieved results is limited. 
}\label{table:kcg_ablation}
\vspace{-0.4cm}
\end{table}

\noindent\textbf{Result on Poetry Appreciation.}
As shown in Table \ref{table:prom}, our method demonstrates effectiveness on Chinese classical poetry appreciation. When built upon the Llama-3.1-8B-Instruct base model, our framework achieves 60.55\% basic-level matching (+38.5 points over vanilla Llama-3.1-8B) and 64.97\% advanced-level accuracy, surpassing all closed-source models in basic-level adaptation (vs Qwen-Plus-FS=56.78) while maintaining competitive intermediate-level performance (80.07 vs 88.01). Notably, the advanced-level result (64.97\%) approaches closed-source models' upper bound (Claude-3.5-FS=61.19, Qwen-Plus-FS=61.85), proving our CLAF effectively handles cultural domain-specific complexity despite the base model's limited Chinese poetry training data.

\vspace{-0.3cm}
\subsection{Ablation Study}

\noindent\textbf{Ablation Study in CLAF.} We conducted an ablation study on FK scores and Level Match accuracy. As shown in Table \ref{table:ablation}, removing any component significantly reduces performance. Excluding CAR results in a 16.4\% drop in advanced complexity metrics, emphasizing its importance in maintaining professional depth. Disabling ALSO leads to a 37\% reduction in Basic level match accuracy, highlighting its crucial role in cognitive alignment. 

The KCG module strengthens our framework by guiding generation through keywords from retrieved content. Its impact is most notable when retrieval quality is low.
As shown in Table~\ref{table:kcg_ablation}, KCG significantly boosts performance in the 0–25\% retrieval quality range (\textit{e.g.}, LM-Bas. +3.35, LM-Adv. +6.62) and shows consistent gains in the 25–50\% range (\textit{e.g.}, FK-Adv. +1.09, LM-Bas. +2.98).
These results highlight KCG’s compensatory role, helping maintain generation quality by extracting key concepts from suboptimal retrievals, thereby enhancing the CLAF’s robustness across varying retrieval conditions.

The Figure~\ref{data_ablation} (a) shows the number of knowledge retrieved by CAR at different levels, demonstrating the wider range of knowledge that our CAR can provide as the user's cognition improves.

\noindent\textbf{Ablation Study in Scale.} The scaling experiment in Figure~\ref{data_ablation} (b) reveals two phases. As training data increases from 10\% to 70\% (52→364 samples), accuracy jumps 32.5 points (53.1\%→85.62\%), reaching 95.3\% of full-data performance. Beyond 70\%, gains taper off (70\%→90\%: +3.4; 90\%→100\%: +0.28), indicating performance saturation. This non-linear trend shows our method effectively captures core stylistic features with limited data, while the rest mainly refines edge cases. The plateau after 90\% confirms our 593-sample dataset achieves near-optimal utility via the CAR and KCG.

\subsection{Human Evaluation}

To complement automatic evaluation metrics, we conducted comprehensive human evaluations involving expert assessments and user preference:

\begin{itemize}[itemsep=0pt,topsep=0pt,parsep=0pt,leftmargin=*]
    \item \textbf{Expert Evaluation}: Three graduate-level education specialists assessed 100 tri-level responses using a 5-point Likert scale. Evaluation criteria included fluency (Flu.), cognitive-level alignment (Align.), and pedagogical effectiveness (Guid.).
    \item \textbf{User Preference Evaluation}: We recruited three elementary school students (Bas.) and three graduate students (Adv.) in biochemistry to represent novice and advanced users. Participants compared answers generated by CLAF, ChatGPT-4o, and a supervised fine-tuned (SFT) model, and selected their preferred responses.
\end{itemize}

The results are shown in Table~\ref{tab:human-eval}, and the results show that CLAF has better results than other models in manual evaluation.

\begin{table}[t]
\centering
\resizebox{\linewidth}{!}{
\begin{tabular}{lccccc}
\toprule
{Model} & {Flu.} & {Align.} & {Guid.} & {Bas.} & {Adv.} \\
\midrule
CLAF(our)      & 4.41 & \textbf{3.89} & \textbf{4.58} & \textbf{53.47\%} & \textbf{60.34\%} \\
ChatGPT-4o & \textbf{4.52} & 3.51 & 4.23 & 32.15\% & 21.52\% \\
SFT        & 4.23 & 3.47 & 4.31 & 14.38\% & 18.14\% \\
\bottomrule
\end{tabular}
}
\caption{\textbf{Human Evaluation Results.} Human evaluation indicate that outputs from CLAF are more preferred compared to other open-source models, demonstrating both the model quality and the effectiveness of Scale.}
\vspace{-0.5cm}
\label{tab:human-eval}
\end{table}

\vspace{-0.1cm}
\subsection{Case Study}

Figure~\ref{case} showcases CLAF’s hierarchical advantages across three comparative levels.
At the basic level, our CLAF module enables vivid metaphors and anthropomorphic language (\textit{e.g.}, ``sky dancers''), outperforming GPT-4o's more technical phrasing (\textit{e.g.}, ``buoyancy'').
At the intermediate level, both models explain density differences, but ours adds clarity with a structured three-step logic: ``Archimedes'' law → density comparison → force chain.”
At the advanced level, both give accurate answers, but our CAR module distinguishes itself with precise mathematical expressions (\textit{e.g.}, $F_b=\rho Vg$, $H_2=0.08988\ \text{kg/m}^3$), offering a solid foundation for academic research.

\vspace{-0.2cm}
\section{Conclusions}
\vspace{-0.1cm}
We address the challenge of cognitive-level misalignment in LLM-based generation by introducing Scale, a dataset with tri-level cognitively aligned answers. Building on Scale, we propose CLAF, a modular framework that adapts content and style to users’ cognitive capacity. Experiments show that CLAF significantly improves cognitive alignment and controllability. This work lays the groundwork for cognitively adaptive generation across education and other user-facing domains.

\noindent\textbf{Acknowledge:}This research was supported by grants from the “Pioneer” and “Leading Goose” R\&D Program of Zhejiang (2025C02022), National Natural Science Foundation of China (No.62307032) and the Key Research and Development Program of Zhejiang Province (No.2024C03270).

\section*{Limitations}
In this section, we discuss the limitations of our work as below:
\begin{itemize}
\item Currently, CLAF categorizes users into three levels. While this approach provides a general framework, a more refined categorization might lead to better adaptive responses and more accurate modeling of student learning needs. We will leave this as future work.
\item Our work focuses on cognitive alignment and acknowledges the limits of a three-level categorization; however, it does not account for motivational, affective, or individual user differences, which may affect its practical applicability in real-world settings. Future extensions will explore incorporating these factors into the framework.
\end{itemize}

\bibliography{custom}

\appendix

\section{Detail of Cognitive Level}
\label{appendix:cognitive_detial}
In our study, we construct a dataset by carefully selecting a diverse set of questions $q$ from each source, along with their corresponding answers $a$. To capture cognitive adaptability, we generate three distinct responses for each answer, each tailored to a specific cognitive level: \textbf{basic}, \textbf{intermediate}, and \textbf{advanced}. These levels correspond to key learning stages, ensuring that the responses are both educationally relevant and cognitively engaging:

\begin{itemize}[itemsep=0pt,topsep=0pt,parsep=0pt,leftmargin=*]
    \item \textbf{Basic level}: This level targets early childhood to elementary school users. It focuses on providing simplified explanations and structured guidance to support foundational understanding and cognitive growth.
    \item \textbf{Intermediate level}: Geared towards middle and high school students, this level introduces more complex concepts and encourages moderate reasoning. It aims to bridge the gap between basic comprehension and advanced analytical skills, fostering critical thinking and problem-solving abilities.
    \item \textbf{Advanced level}: Designed for undergraduate students and beyond, this level explores complex, abstract concepts that require strong analytical skills. It challenges users to engage with sophisticated ideas, promoting deep understanding and intellectual development.
\end{itemize}

\section{Dataset Curation Detail}
\label{appendix:data_detail}
\subsection{Dataset Matadata}
\begin{table}[h]
\centering

\begin{tabular}{lc}
\cline{1-2}
Category & Quantity \\ \cline{1-2}
science & 153 \\
nature & 140 \\
biology & 192 \\
cosmology & 35 \\
poetry & 73 \\ 
\hline
total  & 593 \\
\cline{1-2}
\end{tabular}
\end{table}

\subsection{Detail about Termin ology Adaptation}
This appendix documents the prompt design and methodological details for terminology processing in our tiered knowledge adaptation framework.

\subsubsection{Terminology Extraction}
We ultimately obtained 1985 pairs belonging to the mapping using the following prompts.

Identify domain-specific terms requiring adaptation.
\noindent \textbf{Prompt Template:}

\begin{mdframed}[
    backgroundcolor=red!10, 
    linecolor=black,        
    innerleftmargin=5pt,     
    innerrightmargin=5pt,    
    innertopmargin=5pt,      
    innerbottommargin=5pt,   
    skipabove=10pt,          
    skipbelow=10pt           
]

{\small
\begin{Verbatim}[breaklines=true]
You are a linguistics expert analyzing educational content. Carefully extract all key technical terms from the following text that require complexity adjustment for different learner levels. Consider:

1. Specialized vocabulary beyond daily usage
2. Abstract conceptual terminology
3. Domain-specific jargon
4. Terms with complexity variations across cognitive levels
Return a JSON list without commentary:
{
  "terms": ["term1", "term2", ...]
}
Text: {insert_content}
\end{Verbatim}
}
\end{mdframed}

\subsubsection{Cognitive-Level Mapping Prompt}

Generate tiered synonyms aligned with educational stages:

\noindent \textbf{Prompt Template:}

\begin{mdframed}[
    backgroundcolor=red!10, 
    linecolor=black,         
    innerleftmargin=5pt,     
    innerrightmargin=5pt,    
    innertopmargin=5pt,      
    innerbottommargin=5pt,   
    skipabove=10pt,          
    skipbelow=10pt           
]
{\small
\begin{Verbatim}[breaklines=true]
As an expert lexicographer with pedagogical training, generate three cognitive-level appropriate synonyms for the term "{term}" using these guidelines:

Basic ({target_age}):  
- Simple concrete language  
- Maximum 2 syllables preferred  
- Use everyday analogues  
Example: "Photosynthesis" → "Plant food-making"
Intermediate ({target_age}):  
- Introduce conceptual components  
- Allow 3-4 syllables  
- Maintain precision while improving accessibility  
Example: "Mitochondria" → "Cell energy factories"
Advanced ({target_age}):  
- Technical precision prioritized  
- Permit specialized jargon  
- Match academic literature usage  
Example: "Catalyst" → "Chemical reaction mediator"

Provide JSON output:  
{
  "term": "{term}",
  "cognitive_mapping": {
    "basic": "...",
    "intermediate": "...", 
    "advanced": "..."  
  }
}
\end{Verbatim}
}
\end{mdframed}

\subsection{Detail about Syntactic Adaptation}
\label{appendix:syntactic}

\subsubsection{BNF Constraints for Cognitive Levels}
We formalize syntactic complexity control through BNF grammars:

\noindent \textit{Basic Level Grammar}

\begin{mdframed}[
    backgroundcolor=red!10,
    linecolor=black,        
    innerleftmargin=5pt,     
    innerrightmargin=5pt,    
    innertopmargin=5pt,      
    innerbottommargin=5pt,  
    skipabove=10pt,          
    skipbelow=10pt           
]

{\small
\begin{Verbatim}[breaklines=true]
<S> ::= <SimpleNounPhrase> <PresentTenseVerb> <Object>  
<SimpleNounPhrase> ::= [Determiner] [Adjective] Noun  
<Object> ::= Noun | "that" <SimpleClause>  
<SimpleClause> ::= <SimpleNounPhrase> Verb  
\end{Verbatim}
}
\end{mdframed}

Features:
Only simple present tense
Maximum 1 subordinate clause
Prohibited structures: passives, modals, gerunds

\noindent \textit{Intermediate Level Grammar}

\begin{mdframed}[
    backgroundcolor=red!10, 
    linecolor=black,         
    innerleftmargin=5pt,     
    innerrightmargin=5pt,    
    innertopmargin=5pt,      
    innerbottommargin=5pt,   
    skipabove=10pt,          
    skipbelow=10pt           
]

{\small
\begin{Verbatim}[breaklines=true]
<S> ::= <ComplexNounPhrase> <VerbPhrase> [Conjunction <S>]  
<VerbPhrase> ::= [Modal] [Adverb] Verb [PrepositionalPhrase]  
<ComplexNounPhrase> ::= [Determiner] [Adjective+] Noun [RelativeClause]  
<RelativeClause> ::= "that" <VerbPhrase> | "which" <VerbPhrase>  
\end{Verbatim}

}
\end{mdframed}
Features:
Allows past/future tenses
Permits 2-level clause nesting
Limited modals (can/may/will)

\noindent \textit{Advanced Level Grammar}

\begin{mdframed}[
    backgroundcolor=red!10, 
    linecolor=black,         
    innerleftmargin=5pt,     
    innerrightmargin=5pt,    
    innertopmargin=5pt,      
    innerbottommargin=5pt,   
    skipabove=10pt,          
    skipbelow=10pt           
]

{\small
\begin{Verbatim}[breaklines=true]
<S> ::= <Nominalization> | <PassiveVoice> | <Conditional>  
<PassiveVoice> ::= <NounPhrase> "is" VerbPastParticiple [PrepositionalPhrase]  
<Conditional> ::= "If" <S> "," ("then" <S> | <ModalVerb> <S>)  
<Nominalization> ::= <GerundPhrase> Verb <ComplexNounPhrase>  
\end{Verbatim}

}

\end{mdframed}

Features:
Supports all verb forms (gerunds, participles)
Allows multi-clause embeddings
Permits abstract syntactic constructions

\subsubsection{Syntax Adjustment Prompt}

Transform text to match target cognitive level's BNF grammar:

\noindent \textbf{Prompt Template:}

\begin{mdframed}[
    backgroundcolor=red!10, 
    linecolor=black,        
    innerleftmargin=5pt,    
    innerrightmargin=5pt,   
    innertopmargin=5pt,     
    innerbottommargin=5pt, 
    skipabove=10pt,        
    skipbelow=10pt        
]

{\small
\begin{Verbatim}[breaklines=true]
As a linguistic editor, rewrite the following text strictly adhering to these BNF constraints for {cognitive_level}:  
{insert_relevant_BNF_rules}  
Key requirements:  
1. Sentence structure must validate against BNF  
2. Lexical complexity matches {cognitive_level} terminology  
3. Preserve original semantic content  

Input: {text}  
Output (JSON):  
{
  "original": "...",
  "restructured": "...", 
  "validation": {"pass": bool, "issues": []}
}  
\end{Verbatim}

}

\end{mdframed}

\subsubsection{Consistency Revision Pipeline}
Embeddings are generated using \texttt{Sentence-BERT} (all-mpnet-base-v2) with cosine similarity measurement.

\noindent \textbf{Prompt Template:}

\begin{mdframed}[
    backgroundcolor=red!10,
    linecolor=black,         
    innerleftmargin=5pt,     
    innerrightmargin=5pt,    
    innertopmargin=5pt,     
    innerbottommargin=5pt,  
    skipabove=10pt,         
    skipbelow=10pt          
]

{\small
\begin{Verbatim}[breaklines=true]
Input:  
```json
{
  "basic": "[simple sentence]",
  "intermediate": "[mid-level sentence]", 
  "advanced": "[complex sentence]"
}
Instruction:
"Detect factual conflicts across three cognitive-level sentences. Revise only conflicting parts using strikethrough→correction while preserving original complexity:
Cross-check scientific accuracy
Modify contradictions only
Maintain sentence structure
Output:
{
  "revisions": {
    "basic": "[revised]",
    "intermediate": "[revised]",
    "advanced": "[revised]"
  },
}
\end{Verbatim}

}

\end{mdframed}

\section{Construction of Adaptive Knowledge Graph}
\label{appendix:rag_build}
\subsection{Extraction of Concepts}

Extracting surrogate cognitive-level entities and relations from text

\noindent \textbf{Prompt Template:}

\begin{mdframed}[
    backgroundcolor=red!10, 
    linecolor=black,         
    innerleftmargin=5pt,    
    innerrightmargin=5pt,   
    innertopmargin=5pt,     
    innerbottommargin=5pt,  
    skipabove=10pt,          
    skipbelow=10pt          
]

{\small
\begin{Verbatim}[breaklines=true]
Given a text document that is potentially relevant to this activity and a list of entity types, identify all entities of those types from the text and all relationships among the identified entities. Use {language} as output language.
-Steps-
1. Identify all entities. For each identified entity, extract the following information:
- entity_name: Name of the entity, use same language as input text. If English, capitalized the name.
- entity_type: One of the following types: [{entity_types}]
- entity_description: Comprehensive description of the entity's attributes and activities
- entity_cognitiev_level: One of the following cognitive levels:
    Basic level: This level targets early childhood to elementary school learners. It focuses on providing simplified explanations and structured guidance to support foundational understanding and cognitive growth.
    Intermediate level: Geared towards middle and high school students, this level introduces more complex concepts and encourages moderate reasoning. It aims to bridge the gap between basic comprehension and advanced analytical skills, fostering critical thinking and problem-solving abilities.
    Advanced level: Designed for undergraduate students and beyond, this level explores complex, abstract concepts that require strong analytical skills. It challenges learners to engage with sophisticated ideas, promoting deep understanding and intellectual development.
  \end{itemize}
Format each entity as ("entity"⟨entity_name⟩ ⟨entity_type⟩ ⟨entity_description⟩ ⟨entity_cognitiev_level⟩)
2. From the entities identified in step 1, identify all pairs of (source_entity, target_entity) that are *clearly related* to each other. 
For each pair of related entities, extract the following information:
- source_entity: name of the source entity, as identified in step 1
- target_entity: name of the target entity, as identified in step 1
- relationship_description: explanation as to why you think the source entity and the target entity are related to each other
- relationship_strength: a numeric score indicating strength of the relationship between the source entity and target entity
- relationship_keywords: one or more high-level key words that summarize the overarching nature of the relationship, focusing on concepts or themes rather than specific details
- relationship_cognitiev_level: A cognitive level indicating the cognitiev_level required to understand the relationship. This should be calculated by considering both the complexity of the relationship itself and the average cognitiev_level of the source and target entities. Use the following guideline:
    - Calculate the average cognitiev_level of the source and target entities.
    - Consider the inherent complexity of the relationship.
    - Assign a cognitiev_level level from the three cognitive levels.
Format each relationship as ("relationship" ⟨source_entity⟩ ⟨target_entity⟩ ⟨relationship_description⟩ ⟨relationship_keywords⟩ ⟨relationship_strength⟩ ⟨relationship_cognitiev_level⟩)
3. Identify high-level key words that summarize the main concepts, themes, or topics of the entire text. These should capture the overarching ideas present in the document.
Format the content-level key words as ("content_keywords"⟨high_level_keywords⟩)
4. Return output in {language} as a single list of all the entities and relationships identified in steps 1 and 2. Use **{record_delimiter}** as the list delimiter.
5. When finished, output {completion_delimiter}

\end{Verbatim}

}
\end{mdframed}

\subsection{Multi-layer Knowledge Graph Construction Algorithm}
\label{algo:hie}

\begin{algorithm}
\caption{Adaptive Retrieval Graph Construction}
\begin{algorithmic}[1]
\REQUIRE Text corpus $D$, maximum level $L$
\ENSURE Adaptive Retrieval Graph $G_{total}$
\STATE $\mathcal{E},\mathcal{R} \leftarrow \text{Extration}(D)$
\FOR{each $e_i \in \mathcal{E}$}
    \STATE $l_i \leftarrow \text{HierarchyAssigner}(e_i)$
\ENDFOR
\FOR{each $(e_i, r, e_j) \in (\mathcal{E},\mathcal{R})$}
    \IF{ $l_i = l_j$}
        \STATE $G_{total}.\text{add\_edge}(e_i, e_j, r)$
    \ENDIF
    \IF{$|l_i - l_j| \leq 1$ \textbf{and} $l_i \neq l_j$}
    
        \STATE $G_{total}.\text{add\_crosslink}(e_i, e_j)$
    \ENDIF
\ENDFOR
\RETURN $G_{total}$
\end{algorithmic}
\end{algorithm}

\subsection{Hierarchical Knowledge Retrieval}
\label{algo:hie2}

\begin{algorithm}
\caption{Hierarchical Knowledge Retrieval}
\begin{algorithmic}[1]
\REQUIRE Query $q$, cognitive level $c \in \{0, 1, 2\}$, graph $G_{total}$, parameters top-k $k$, depth $d$
\ENSURE Knowledge subset $K^{(k)}_c$
\STATE Set maximum level based on cognitive level:
\STATE $l_{max} \leftarrow c$
\STATE $G_c \leftarrow \{e \in G_{total} \mid l_e \leq l_{max}\}$
\STATE $\phi_q \leftarrow \text{QueryRewriter}(q, c)$
\STATE Initialize result set $K^{(k)}_c \leftarrow \emptyset$
\STATE Perform initial query traversal:
\STATE $S_k \leftarrow \text{TopK}(\text{Neighbor}(\phi_q, G_c), k)$
\STATE Add results to $K^{(k)}_c$: $K^{(k)}_c \leftarrow K^{(k)}_c \cup S_k$
\FOR{each $e_i \in S_k$}
    \STATE Retrieve neighbors at depth $d + 1$: $N_i \leftarrow \text{NeighborDepth}(e_i, G_c, d+1)$
    \STATE Add neighbors to $K^{(k)}_c$: $K^{(k)}_c \leftarrow K^{(k)}_c \cup N_i$
\ENDFOR
\RETURN $K^{(k)}_c$
\end{algorithmic}
\end{algorithm}

\section{Implementation Details.}

The experiments were conducted on a cluster with 8×NVIDIA A100 80GB GPUs, utilizing BF16 mixed precision and FlashAttention-2 for computational efficiency. The specific configurations are as follows: (1) Hierarchical Retrieval-Augmented Generation: We extracted a knowledge graph comprising 6,244 entities and 6,364 relations by setting the chunk token size to 600 and the chunk overlap token size to 100; (2) DPO Training: Using Llama3.1-8B-Instruct as the base model, we first performed 1 epoch of Supervised Fine-Tuning (SFT) with a global batch size of 64 (micro batch size of 16) and a learning rate of 5e-6, followed by 1 epoch of DPO with a learning rate of 5e-7 and beta=0.1, maintaining the same batch configuration as in SFT; (3) KCG: We extracted Control Center features from the DPO-trained model and implemented dynamic parameter control.

\section{Metrics}
\label{appendix:metrics}
\noindent \textbf{Prompt Template:}

\begin{mdframed}[
    backgroundcolor=red!10, 
    linecolor=black,        
    innerleftmargin=5pt,    
    innerrightmargin=5pt,  
    innertopmargin=5pt,     
    innerbottommargin=5pt,  
    skipabove=10pt,         
    skipbelow=10pt          
]

{\small
\begin{Verbatim}[breaklines=true]
You are tasked with evaluating how well a given response matches the intended audience level. Consider the following audience types and their criteria:


Basic level: This level targets early childhood to elementary school learners. It focuses on providing simplified explanations and structured guidance to support foundational understanding and cognitive growth.
Intermediate level: Geared towards middle and high school students, this level introduces more complex concepts and encourages moderate reasoning. It aims to bridge the gap between basic comprehension and advanced analytical skills, fostering critical thinking and problem-solving abilities.
Advanced level: Designed for undergraduate students and beyond, this level explores complex, abstract concepts that require strong analytical skills. It challenges learners to engage with sophisticated ideas, promoting deep understanding and intellectual development.

Audience Type 0 (Basic level):
Is the response fun and engaging?
Does it use simple knowledge points and avoid complex vocabulary?
Are analogies and metaphors used appropriately?
Is any difficult content explained in simple terms?

Audience Type 1 (Intermediate level):
Does the response provide normal knowledge points?
Is it based on common sense and easily understandable?
Audience Type 2 (Advanced level):
Is the response professional and detailed?
Does it use technical language appropriate for experts in the field?
Question: {question}

Answer for Audience Type 0: {answer_type_0}

Answer for Audience Type 1: {answer_type_1}

Answer for Audience Type 2: {answer_type_2}

Evaluate each response based on how well it aligns with the specified audience type and provide a score out of 100 for each. Output only the scores as numbers.
\end{Verbatim}

}
\end{mdframed}

\section{Dataset Construction Validation}
\subsection{Terminology Mapping Validation}
\label{appendix:dataset_val}
\begin{itemize}[itemsep=0pt,topsep=0pt,parsep=0pt,leftmargin=*]
\item \textbf{Expert Panel}: Three biochemistry graduate researchers assessed the conceptual accuracy and complexity of mapped terms.
\item  \textbf{Consensus Mechanism}: Terms were refined collaboratively if inconsistencies were flagged.
\item  \textbf{User Testing}: Clarity and accessibility were validated through user feedback.
\end{itemize}

\subsection{Educational Suitability Review}

\begin{itemize}[itemsep=0pt,topsep=0pt,parsep=0pt,leftmargin=*]
\item  \textbf{Panel}: Five education specialists in curriculum design assessed responses across basic, intermediate, and advanced tiers.
\item  \textbf{Criteria}: Each response was graded for age-appropriateness and conceptual clarity.
\item  \textbf{Revision}: 278 out of 593 samples were iteratively improved (46.8\% revision rate).

\end{itemize}

\section{Details of Human Assessment}

\label{appendix:human_assess}
\subsection{Expert Evaluation}
Three graduate-level education specialists evaluated responses across all three difficulty levels using a 5-point Likert scale, where a higher score indicates better performance. They assessed:
\begin{itemize}[itemsep=0pt,topsep=0pt,parsep=0pt,leftmargin=*]
    \item \textbf{Fluency}: Language clarity and logical structure.
    \item \textbf{Cognitive Level Alignment}: Appropriateness for target users.
    \item \textbf{Pedagogical Guidance}: Educational effectiveness.
\end{itemize}
A total of 100 questions were evaluated.
 
\subsection{Users Evaluation}
To assess real-world effectiveness, we recruited:
\begin{itemize}[itemsep=0pt,topsep=0pt,parsep=0pt,leftmargin=*]
    \item \textbf{Basic Level} Three elementary school students.
    \item \textbf{Advanced Level} Three graduate students.
\end{itemize}
Participants compared responses from three systems (CLAF, ChatGPT-4o, and an SFT-tuned model) and selected their preferred answers.

\section{Detail of Personnel and Computational Cost}
\label{appendix:cost}
\subsection{Human Resource Compensation}

To support the evaluation and dataset construction processes, we involved several participants from both education and biochemistry domains. All participants were compensated accordingly. For international clarity, compensation amounts are approximated in U.S. dollars.

During the expert evaluation phase, we recruited three graduate-level education specialists. Each specialist was compensated approximately \$60 for one and a half days of participation.

In the users evaluation stage, we involved three elementary school students and three graduate students in biochemistry. The graduate students received approximately \$30 each, while the elementary students were compensated about \$20 per participant, all for one day of participation.

During the dataset construction process, three biochemistry graduate students helped validate the terminology mapping. Each received around \$40 for two days of work.

Finally, five graduate students specializing in education were recruited for the final dataset review. Each participant was compensated approximately \$40 for two days of evaluation work.

\subsection{Computational Efficiency}

Despite the layered architecture of our framework, the system remains computationally efficient and scalable.

The construction of the entire domain-specific knowledge graph, consisting of over 6,000 entities and relations, took less than ten minutes and cost approximately \$1 using a commercially available API. This demonstrates the feasibility of rapid knowledge graph generation.

CLAF operates with computational requirements similar to those of the LLaMA-3.1-8B model, requiring around 20GB of VRAM. Further optimization through quantization can reduce these requirements, enabling deployment on standard hardware.

The CAR multi-step retrieval module is highly efficient, achieving an average retrieval time of approximately 2 seconds per query, which supports real-time interactive use without noticeable latency for end-users.

These cost and performance metrics confirm that our framework can be practically adopted in large-scale educational settings without significant human or computational overhead.

\end{document}